\documentclass[runningheads]{llncs}
\usepackage[T1]{fontenc}
\usepackage{graphicx,verbatim}
\usepackage{hyperref}
\usepackage{color}
\usepackage{lipsum}
\usepackage{amsmath}
\usepackage{booktabs}
\usepackage{subcaption}
\usepackage{bm}
\usepackage{xcolor}
\usepackage{enumitem}
\usepackage[linesnumbered,ruled]{algorithm2e}
\graphicspath{{./fig/}}

\begin{document}
\title{Parametric shape models for vessels learned from segmentations via differentiable voxelization}
\titlerunning{Parametric shape models for vessels}
\author{
    \mbox{Alina F. Dima} \inst{1} \and
    \mbox{Suprosanna Shit} \inst{2} \and
    \mbox{Huaqi Qiu} \inst{1} \and
    \mbox{Robbie Holland} \inst{3} \and
    \mbox{Tamara T. Mueller} \inst{1} \and
    % \mbox{Fabio Antonio Musio} \inst{2} \and
    \mbox{Fabio Musio} \inst{2,4} \and
    \mbox{Kaiyuan Yang} \inst{2} \and
    \mbox{Bjoern Menze} \inst{2} \and
    \mbox{Rickmer Braren} \inst{5,6} \and
    \mbox{Marcus Makowski} \inst{5} \and
    \mbox{Daniel Rueckert} \inst{1, 6, 7, 8}
}
\institute{
Chair for AI in Healthcare and Medicine, Technical University of Munich (TUM) and TUM University Hospital, Munich, Germany
\and
University of Zurich, Zurich, Switzerland
\and
Stanford Center for Artificial Intelligence in Medicine and Imaging, Stanford University, USA
\and
Center for Computational Health, ZHAW, Zurich, Switzerland
\and
Institute for Diagnostic and Interventional Radiology, TUM University Hospital, Munich, Germany
\and
German Cancer Consortium (DKTK), Munich partner site, Heidelberg, Germany
\and
Department of Computing, Imperial College London, UK
\and
Munich Center for Machine Learning (MCML), Munich, Germany
\\
\email{alina.dima@tum.de}
}
\authorrunning{A. Dima et al.}
\maketitle
\begin{abstract}

Vessels are complex structures in the body that have been studied extensively in multiple representations.
While voxelization is the most common of them, meshes and parametric models are critical in various applications due to their desirable properties.
However, these representations are typically extracted through segmentations and used disjointly from each other.
We propose a framework that joins the three representations under differentiable transformations.
By leveraging differentiable voxelization, we automatically extract a parametric shape model of the vessels through shape-to-segmentation fitting, where we learn shape parameters from segmentations without the explicit need for ground-truth shape parameters.
The vessel is parametrized as centerlines and radii using cubic B-splines, ensuring smoothness and continuity by construction.
Meshes are differentiably extracted from the learned shape parameters,
resulting in high-fidelity meshes that can be manipulated post-fit.
Our method can accurately capture the geometry of complex vessels, as demonstrated by the volumetric fits in experiments on aortas, aneurysms, and brain vessels.
%
% Keywords
\end{abstract}

\section{Introduction}

Vessels, the transport network of the circulatory system, are ubiquitous in the body.
Impaired vessel function has profound implications in cardiovascular disease, leading to stroke or heart failure.
In medical imaging, vessel analysis is multifaceted, involving vessel segmentation\cite{aortaseg24}, flow simulations~\cite{numata2016blood}, curved planar reformation~\cite{rist2023flexible}, and stricture detection via cross-sectional analysis~\cite{de2022quantification}.

Segmentation is often the starting point of vessel analysis, owing to its versatility and high-performance algorithms~\cite{isensee2024nnu}.
However, it \emph{lacks semantic information}.
Flow analysis relies on high-quality meshes, whereas cross-sectional analysis relies on centerline and radius estimates.
Marching cubes~\cite{lorensen1998marching} is a popular method of extracting meshes from segmentations, but the mesh properties are challenging to manipulate, requiring laborious postprocessing.
Implicit representations can extract high-fidelity surfaces~\cite{alblas2023implicit}, but the lack of control over the mesh structure persists.
Data-driven methods~\cite{timmins2022deep} come with higher data requirements and ground truth that matches the number of mesh elements.
Centerline extraction is often performed automatically using skeletonization algorithms\cite{menten2023skeletonization,voreen}, resulting in discrete, noisy centerlines.
% Attempts have been made to learn centerlines from 2D projections~\cite{kozinski2020tracing}, relying on ground truth annotations.
We propose to merge these representations in an explicit \textbf{parametric shape model}.
Our model consists of smooth and continuous centerlines and cross-sectional radii modelled by splines, resulting in a representation that is \emph{sparse by construction}.
From there, we derive both mesh and volumetric representations using \emph{differentiable voxelization} to enable gradient-based model fitting.

% Graphical abstract
\begin{figure}[t!]
    \centering
    \includegraphics[width=\textwidth,height=0.22\textheight]{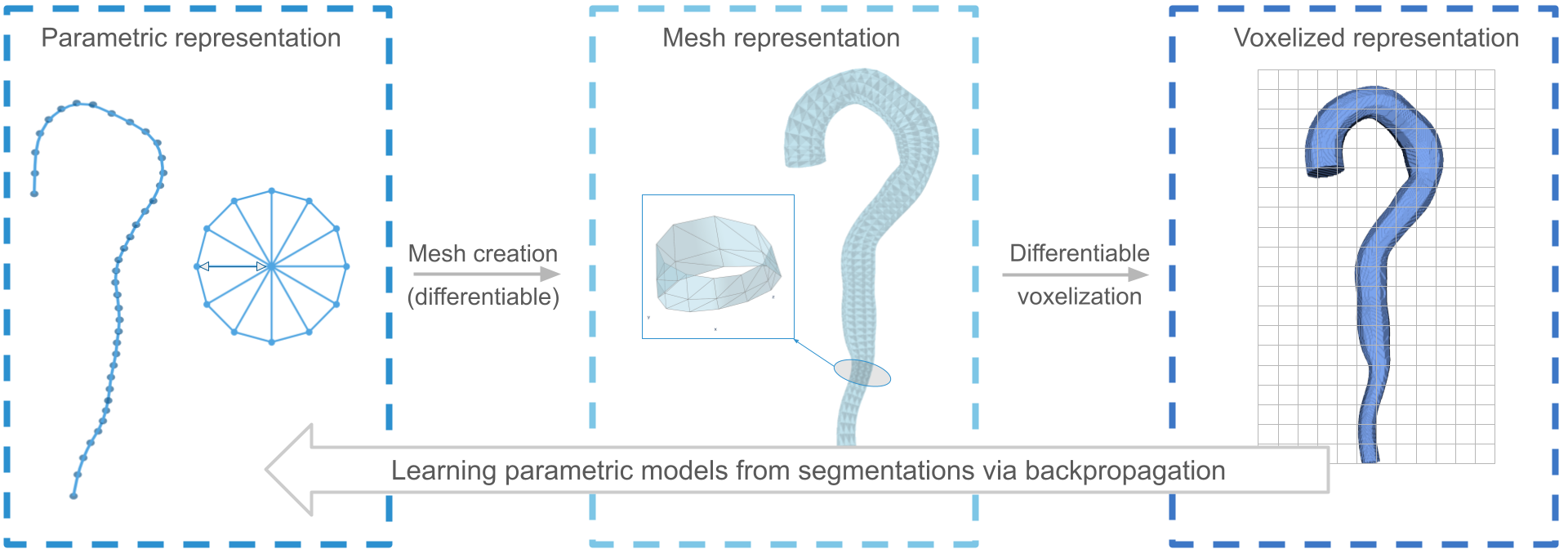}
    \caption{
    We introduce parametric vessel models that can be learned from segmentations using differentiable voxelization.
    Our formulation allows for a seamless conversion between three representation spaces: parametric $\rightarrow$ mesh $\rightarrow$ voxelized.
    }%
    \label{fig:graphical_abstract}
\end{figure}

Differentiable voxelization in medical imaging is appealing for leveraging meshes and segmentations jointly.
A projection-based approximate voxelization has been proposed for cardiac shape fitting~\cite{meng2023deepmesh}, while another approximate differentiable voxelization based on thin plate spline warping has been employed for the left ventricles, both of which have restricted applicability to other shapes.
In the 2D domain, instance segmentation~\cite{lazarow2022instance,bransby2024polycore} has benefited from differentiable rasterization, the 2D counterpart of voxelization.
In computer graphics, voxelization algorithms are commonplace~\cite{schwarz2010fast}; however, as far as we know, no such algorithms have been employed in learning-based 3D medical imaging, potentially due to the prohibitive memory cost of the gradients.
Our work leverages a differentiable voxelization algorithm for converting meshes to voxelizations in an \emph{end-to-end differentiable} manner.
This allows us to \textbf{fit vessel parameterizations} using supervision from \emph{segmentations} via intermediate mesh representations.
We propose to learn parametric shape models of isolated vessels fitted to segmentations without requiring parametric ground truth.
Using a differentiable voxelization loss, our parametric models allow for seamless and differentiable conversion between centerlines and radii, meshes, and segmentations.
We fit these models per vessel, using a preliminary centerline, vessel endpoints, and segmentation as supervision.
The parameters can be \textbf{adjusted post-fitting}, which is desirable for surgical applications.
Our contributions can be summarized as follows:
\\
1. We introduce \textbf{parametric vessel models} that can be fitted to a reference segmentation \emph{without parametric ground truth} end-to-end.
The use of differentiable voxelization enables a seamless conversion between three representations: \textbf{shape models}, \textbf{meshes}, and \textbf{segmentations}, enabling joint manipulation and gradient-based optimization.
\\
2. We utilize a neural network to fit these models from images, leveraging rich visual information while exploiting the regularization properties of the network. The parametric models are based on cubic B-splines, which are continuous and sparse by design.
\\
3. We provide a training recipe and implementation for fitting parametric vessel shape models to segmentations, including a \textbf{differentiable voxelization loss} for 3D voxel-based supervision and a \textbf{four-stage training process} that addresses the gradient conflicts inherent to joint parameter optimization.
\\
4. We demonstrate that our parametric models enforce \emph{sparseness, interpretability, biomechanical plausibility}, and \emph{connectivity}, while maintaining the high fitting accuracy of SOTA segmentation methods, achieving the best of both worlds.
\\

% %%%%%%%%%%%%%%%%%%%%%%%%%%%%%%%%%%%%%%%%%%%%%%%%%%%%%%%%
\section{Method}~\label{sec:method}
\begin{figure}[t!]
	\centering
	\includegraphics[width=\textwidth]{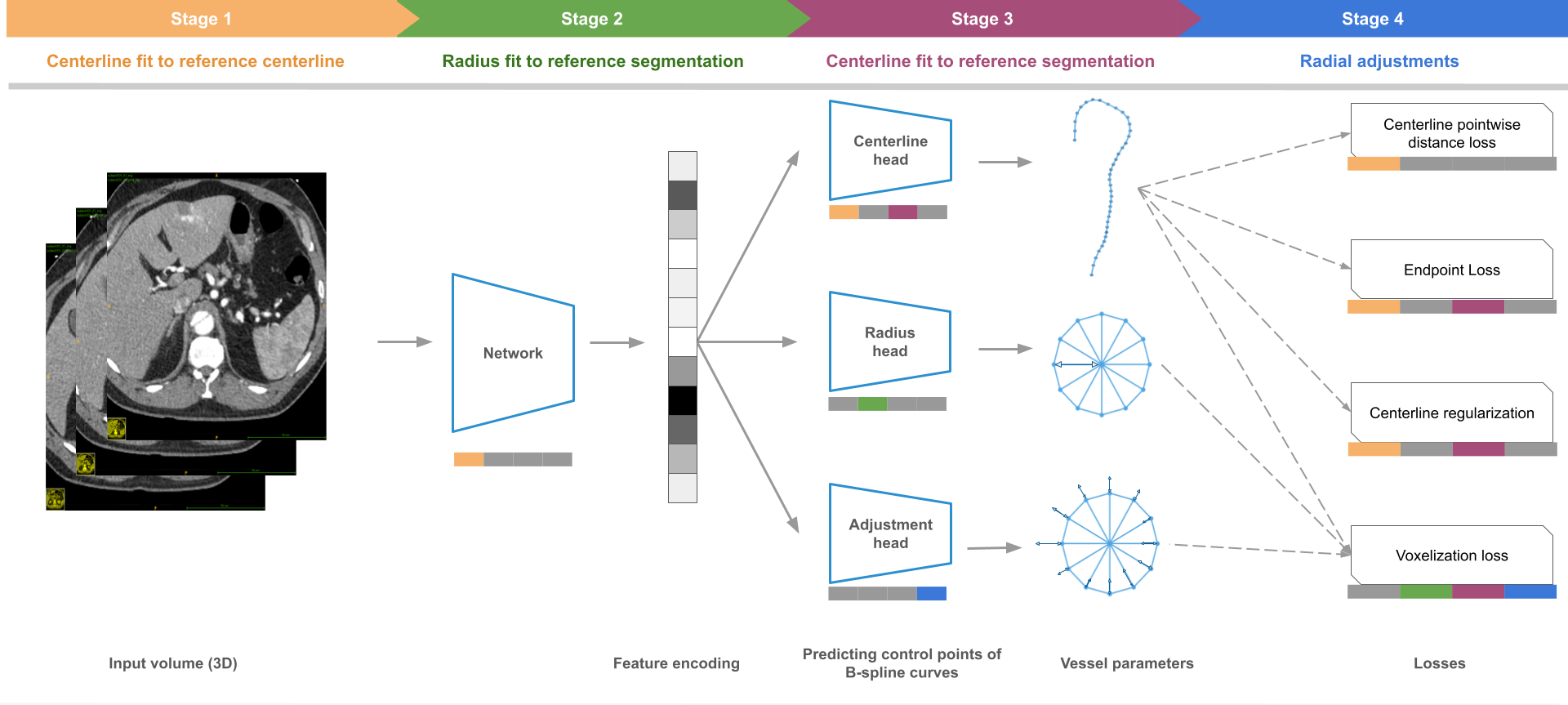}
	\caption{
    An overview of the network components and training stages (Section~\ref{sec:training_stages}).
    In each of the four stages (top of the figure, color-coded), only one prediction head is actively trained.
    %  due to the complexity of the optimization problem.
    % The presence of a stage color below a component/loss indicates its use during the corresponding stage.
    % As a backbone network, we use 3D Resnet18.
    % In each of the stages, a different prediction head is individually trained. During the first stage, the backbone and the centerline head are trained using the centerline pointwise distance loss.
    % In the second stage, the radius is learned using the voxelization loss.
    % The centerline is adjusted again in the third stage, this time using the voxelization loss.
    % Finally, the radial adjustments are learned in the fourth stage.
    % Two endpoint loss and centerline regularization are active whenever the centerline is trained.
        }%
	\label{fig:training_stages}%
\end{figure}

% %%%%%%%%%%%%%%%%%%%%%%%%%%%%%%%%%%%%%%%%%%%%%%%%%%%%%%%%
% Overview
We propose an information flow similar to a neural network training scheme~(Figure~\ref{fig:training_stages}).
Starting from an input image, we learn the parametrization of the vessel (centerlines $+$ radii) by training a neural network on a regression task.
% The parametric model is converted to a mesh by defining surface points along discrete radial directions within cross-sections.
The parametric model is converted to a mesh by discretizing the cross-sections along longitudinal and radial directions.
The mesh is then differentiably voxelized to a soft segmentation, on which a segmentation loss is computed based on a reference segmentation.
We use our framework to fit models to segmentations \emph{per-sample}, although this approach could be used in a more general setting.
The resulting optimization problem is complex and could only be solved by a careful training scheme involving multiple types of losses and training stages.
Although we focus on single vessels, our parametric models could be extended to vessel trees by adjusting the meshing step to handle branching.
An overview of the representation trio in our pipeline is shown in Figure~\ref{fig:graphical_abstract}.
% # TODO: Move this to the appropriate subsection
% The resulting optimization problem is complex, and benefits from the regularizing properties of neural networks~\cite{ulyanov2018deep}.
% The resulting optimization problem is complex and benefits from the regularizing properties of neural networks~\cite{ulyanov2018deep} and training its components in stages (Fig.~\ref{fig:training_stages}).
%
%
%
% %%%%%%%%%%%%%%%%%%%%%%%%%%%%%%%%%%%%%%%%%%%%%%%%%%%%%%%%
\subsection{Vessel Parametrization and Mesh Creation}%
\label{sec:parametrization}%
\indent The centerline is the backbone of the parametrization, defining the cross-sectional planes.
Within each 2D cross-section, the vessel is approximated as a disk with a fixed radius.
Each radial direction within each cross-section can be further adjusted to allow for anomaly fitting.
% By intersecting the
Since the vessel parameters define points on a 3D surface, we can diffierentiably define a triangle mesh by connecting points in adjacent cross-sections.
% Next, a triangle mesh is differentiably derived from the vessel parameters.
\\
% %%%%%%%%%%%%%%%%%%%%%%%%%%%%%%%%%%%%%%%%%%%%%%%%%%%%%%%%
% Define all the math symbols
\newcommand{\cn}{c^n}
\newcommand{\rn}{r^n}
% %%%%%%%%%%%%%%%%%%%%%%%%%%%%%%%%%%%%%%%%%%%%%%%%%%%%%%%%
\subsubsection{B-splines}
We chose cubic B-splines~\cite{rogers2000introduction} to represent the components of the parametric model, owing to their desirable properties: smoothness and local control.
By predicting a smaller number of control points than the desired number of cross-sections, regularization is inherent.
Another effect of modelling the parameters as continuous curves is the decoupling of the number of output network parameters (the control points) from the resolution of the parametrized representation (the number of cross-sections).
This provides representation flexibility, as a different number of centerline points can be sampled for point-to-point prediction -- where the target has a fixed size --  and mesh generation -- where sparsity is desired as it directly impacts speed.

As the centerline represents 3D points, let $N_c$ be number of control points for each dimension, resulting in $3\times N_c$ overall control points.
The radius models positive quantities, and is hence modelled by $1 \times N_r$ control points.
Typically $N_c > N_r$, as there is more variance in the vessel shape than cross-sectional area.
The radial adjustments are $P$-dimensional, where $P$ is the number of vertices of the regular polygon that approximates the cross-sectional disc, resulting in a total of $P \times N_a$ control points.
In practice, we use $N_a = N_r$.

Given the predicted \mbox{B-spline} control points, we obtain the centerlines, radii and adjustments by sampling from their respective B-spline curves.
For simplicity, we employ an equidistant sampling scheme with \( S \) samples.
This results in the discrete centerline
\mbox{\( \mathcal{C}_S = \{ c_1, \ldots, c_S \} \)},
radius
\mbox{\( \mathcal{R}_S = \{ r_1, \ldots, r_S \} \)},
and adjustments
\mbox{\( \mathcal{A}_S = \{ a_1, \ldots, a_S \} \)},
which define $S$ cross-sections.
The matrix formulation of cubic B-splines~\cite{rogers2000introduction} allows for efficient curve evaluation.
\subsubsection{Cross-sections} are defined at each centerline point as normal to the centerline tangents:
$\overrightarrow{t}_n = \overrightarrow{c}_{n+1} - \overrightarrow{c}_n$, $1 \leq n \leq S$.
For each cross-section $\mathcal{C}_n$, the generating vectors $\overrightarrow{v}_{i}^n$, $\overrightarrow{v}_{j}^n$ are a set of two vectors perpendicular to the tangent vectors and each other.
To ensure a smooth transition between cross-sections, the generating vectors are aligned:
$\overrightarrow{v}_{i}^n = \overrightarrow{v}_{i}^{n-1} \times \overrightarrow{t}_n$,
$\overrightarrow{v}_{j}^n = \overrightarrow{t}_{n} \times \overrightarrow{v}_{i}^{n}$;
the first vector is aligned with $v_{i}^0 = \overrightarrow{(0, 0, 1)}$.

\subsubsection{Triangle Meshes} $\mathcal{M} = (\mathcal{N}, \mathcal{F})$ represent the surface of the object defined by the parametric model, where each cross-section is discretized by a regular polygon with $P$ vertices + individual radial adjustments.
This results in $P$ points per cross-section:
$
\overrightarrow{P_k^n} =
    \overrightarrow{c^n} +
    (r^n + a_k^n) \cos \left( \frac{2 \pi k}{P} \right) \overrightarrow{v}_i^n +
    (r^n + a_k^n) \sin \left( \frac{2 \pi k}{P} \right) \overrightarrow{v}_j^n
$.
\noindent
For the mesh triangles, we split each rectangle joining adjacent cross-sections ($P_{k}^n, P_{k}^{n+1}, P_{k+1}^{n+1}, P_{k+1}^{n}$) into two triangles each.
In addition, the top and bottom cross-sections are represented by triangles between the centerline points and neighboring points within those cross-sections.
This results in a sparse triangle mesh consisting of $P \times S + 2$ vertices and $2 \times P \times S$ triangles.
Alternative mesh representations have not been explored in this work, but could be investigated to improve specific aspects of the mesh quality.
We envision using our proposed mesh representation during the training process to fit the vessel parameters, and during inference one could use a different meshing of the continuous 3D object that the vessel parametrization represents.
%
%
% %%%%%%%%%%%%%%%%%%%%%%%%%%%%%%%%%%%%%%%%%%%%%%%%%%%%%%%%
\subsection{Differentiable Voxelization}%
\label{sec:diff_voxelization}
% %%%%%%%%%%%%%%%%%%%%%%%%%%%%%%%%%%%%%%%%%%%%%%%%%%%%%%%%
\subsubsection{Formulation}
Our voxelization loss is based on the differentiable polygonization loss introduced by~\cite{lazarow2022instance}.
For every output voxel at coordinates $x, y, z$, let SDF be the minimum distance between the voxel center and the mesh surface, and the sign indicating whether the voxel is inside or outside the mesh.
Given a watertight triangle mesh \( \mathcal{M} \), we compute a soft voxelization \( \widehat{\mathrm{VOX}}_{soft} \) as:
%
% % Voxelization
\begin{equation}
    \label{eq:soft_vox}
	\widehat{\mathrm{VOX}}_{soft} (x, y, z) =
	\sigma
		\left(
		\frac{\mathrm{SDF}_\mathcal{M}(x, y, z)}{\tau}
		\right)
\end{equation}
The voxelization loss is the Dice loss between \( \widehat{\mathrm{VOX}}_{soft} \) and the reference segmentation. $\sigma$ is the sigmoid function, while $\tau=0.1$ is a hyperparameter that controls the softness of the voxelization.
% \\
\subsubsection{Practical considerations}

\begin{figure}[t!]
	\centering
	\includegraphics[width=\textwidth,keepaspectratio]{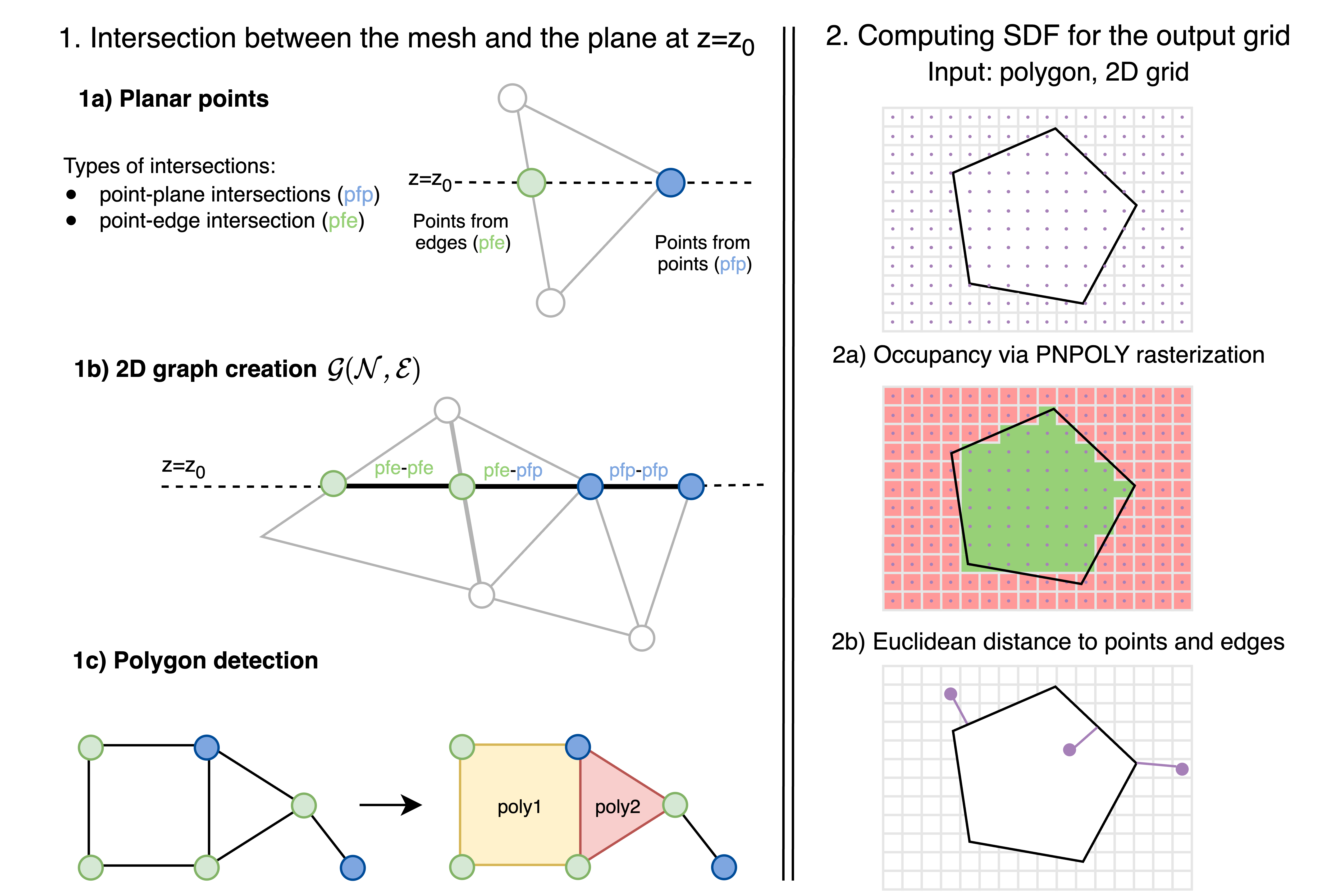}
	\caption{
    Overview of the slice-wise approximation of the SDF.
    A full description of the approach is provided in the main text.
    }
  \label{fig:voxelization_algo}
\end{figure}

Computing the signed distance function (SDF) in a differentiable manner is memory-intensive.
For each voxel in the 3D output space, the signed distance consists of two components: the distance to the nearest mesh element (point, edge or triangle), and its sign (inside or outside the mesh).
To be used as part of a differentiable loss, its gradients also need to be computed and stored, which incurs a significant memory overhead compared to non-differentiable implementations.
With a naive implementation, this results in a large computational graph that does not fit onto a modern 48GB GPU for an average sized 3D image.

Under these constraints, it is necessary to prune the computational graph to reduce memory overhead.
We employ two main strategies: 1. Slice-wise approximation of the SDF, 2. Restricting the output space to a bounding box around the mesh + margin.
The voxels outside the margin are approximated to a soft segmentation of 0, effectively pruning the gradient tree.
The size of the margin is adjusted to the parameter $\tau$; for $\tau=0.1$, the size used was $3$ voxels.

The slice-wise approximation of the SDF is inspired by the approach used by~\cite{lazarow2022instance}, although their approach does not generally extend to 3D.
In order to adapt the method to 3D, we partition the mesh into slices parallel to one of the standard image planes (e.g. XY plane).
An overview of the slice-wise approximation of the SDF is shown in Fig.~\ref{fig:voxelization_algo}.
% In the first step, we partition the mesh into slices parallel to one of the standard image planes (e.g. XY plane).
For each slice, the intersection between the mesh elements and the plane is computed, resulting in a new set of points: some points of the original mesh, and some representing intersections of edges with the plane~(\textbf{1a}).
Two points are connected by an edge if their original structures (point or edge) belong to the same face in the original mesh~(\textbf{1b}).
A graph is defined by the intersection points and edges.
Cycle detection is used to partition the graph into non-overlapping polygons~(\textbf{1c}).
Each polygon is rasterized using the point inclusion in polygon test~\cite{franklinpnpoly} to determine occupancy~(\textbf{2a}), while the distance corresponds to the Euclidean distance to the nearest intersection edge or point~(\textbf{2b}).
By alternating the plane direction between the 3 standard axes in different forward passes, we minimize the error of the approximation.
We provide a PyTorch implementation of the differentiable voxelization in our GitHub repository.
Our implementation is generic and can be used for any 3D mesh, as long as the mesh is watertight.
In its current form, it is computationally expensive due to reduced parallelism, one forward pass taking about 1 minute for the Aorta dataset.
Additional optimizations would be attainable using direct CUDA implementations, which are left for future work.
%
%
%
% %%%%%%%%%%%%%%%%%%%%%%%%%%%%%%%%%%%%%%%%%%%%%%%%%%%%%%%%
\subsection{Losses}%
% %%%%%%%%%%%%%%%%%%%%%%%%%%%%%%%%%%%%%%%%%%%%%%%%%%%%%%%%
%
The backbone of our training is the \emph{voxelization loss}:
\( \bm{\mathcal{L}_{\mathrm{vox}}} = \mathcal{L}_{\text{dice}} ( \widehat{\mathrm{VOX}}_{soft}, V ) \)%
, where \( V \) is the reference segmentation.
Its reliance on the distance to the nearest mesh element requires a good initialization to a preliminary centerline.
We achieve this using a \emph{pointwise distance loss} in the first training stage:
\( \bm{\mathcal{L}_{\mathrm{cl}}} = \sum_{i=1}^{N} \|c_i - \hat{c_i}\|^2 \)%
, where \( c_i \) and \( \hat{c_i} \) are the reference and predicted centerline points, respectively.
We found an additional \emph{endpoint loss} beneficial, as the shape fitting around the boundary is particularly challenging:
\( \bm{\mathcal{L}_{\mathrm{e}}} = \|c_1 - \hat{c_1}\|^2 + \|c_N - \hat{c_N}\|^2 \)%
, where \( c_1 \) and \( c_N \) are the endpoints of the centerline.
\emph{Regularization} based on the centerline curvature variance is also beneficial in smoothing the centerline and preventing degenerate solutions:
\( \bm{\mathcal{L}_{\mathrm{reg}}} = \sum_{i=1}^{N-2} \| \mathrm{c}_i'' \|^2 \).
We train using a linear combination of these losses:
$
\mathcal{L} =
\lambda_{\mathrm{cl}} \mathcal{L}_{\mathrm{cl}} +
\lambda_{\mathrm{e}} \mathcal{L}_{\mathrm{e}} +
\lambda_{\mathrm{vox}} \mathcal{L}_{\mathrm{vox}} +
\lambda_{\mathrm{reg}} \mathcal{L}_{\mathrm{reg}}
$.
The exact values of the weights are determined by the training stages, as described in Section~\ref{sec:training_stages}.
%
%
% %%%%%%%%%%%%%%%%%%%%%%%%%%%%%%%%%%%%%%%%%%%%%%%%%%%%%%%%
\subsection{Training}%
% %%%%%%%%%%%%%%%%%%%%%%%%%%%%%%%%%%%%%%%%%%%%%%%%%%%%%%%%
%
\label{sec:training_stages}
The training process is split into four stages, each focusing on a component of the network (see Figure~\ref{fig:training_stages}).

\begin{itemize}
\item[$\bullet$] \textbf{Stage 1: Preliminary centerline fit}:
In the first stage, we converge to an initial centerline that is automatically extracted from the reference segmentation.
In this stage, we only train the feature extractor and the centerline head.
As loss, we use all the centerline-related losses with the exception of the voxelization loss, as the radius is not yet initialized.
After this stage, the backbone weights are frozen, and the pointwise distance loss is replaced by the voxelization loss.
The weights of the losses follow this principle:
$\lambda_{cl}=1$, $\lambda_e=100$, $\lambda_{vox}=0$, $\lambda_{reg}=10$.
\\
\item[$\bullet$] \textbf{Stage 2: Radius fit}:
Once the centerline is initialized, the radius can be quickly fitted even with a noisy centerline.
In practice, we found the radius fit to converge rapidly and effectively, owing to the inherent regularization of the B-spline representation.
We backpropagate solely through the radius head at this stage, using the voxelization loss, which is the only radius-related loss: $\lambda_{vox}=1$, $\lambda_{reg},\lambda_e,\lambda_{cl}=0$.
\\
\item[$\bullet$] \textbf{Stage 3: Centerline correction}:
Next, we fine-tune the centerline head to correct the centerline based on the voxelization loss, while keeping all other components frozen.
At this stage, we can effectively use the shape matching enabled by the voxelization loss, since the radius has already converged.
The weights of the losses follow the ratios in stage 1, except for the centerline regularization, which has been slightly relaxed: $\lambda_{cl}=0$, $\lambda_e=100$, $\lambda_{vox}=1$, $\lambda_{reg}=5$.
\\
\item[$\bullet$] \textbf{Stage 4: Radial adjustment}:
Finally, after we have fit the radius and centerline, we can tune each radial direction individually to the data.
This stage is important for fitting to anomalous, non-circular cross-sections.
Like in the previous stages, all other components are frozen and the only relevant loss is the voxelization loss.
$\lambda_{vox}=1$, $\lambda_{reg},\lambda_e,\lambda_{cl}=0$.
\end{itemize}
A few key observations motivate the multi-stage approach and its configuration:
\begin{itemize}
    \item
    The SDF-based voxelization loss acts on the object boundary, trying to match the boundary of the target to the nearest boundary.
    If the nearest boundary is on the wrong side of the object, the loss will be unable to escape this local minimum.
    Therefore, in order for the voxelization loss to be effective, the parameters should already be in the vicinity of the target.
    \item
    The centerline and radius components could not be fitted simultaneously; the behaviour we observed in our experiments is quick divergence of both components, even when starting from a good fit.
    We hypothesize that this is due to conflicting gradients not only between the centerline and radius components, but also between different output voxels.
    For any given model configuration, there are multiple conflicting ways of decreasing the voxelization loss, often conflicting between the radius and centerline.
    The same applies to the radial adjustments and the rest of the components.
\end{itemize}
%

% !TEX root = ../main.tex

\section{Experiments and Results}%
\label{sec:experiments_and_results}
% %%%%%%%%%%%%%%%%%%%%%%%%%%%%%%%%%%%%%%%%%%%%%%%%%%%%%%%%
%
%
% Define dataset symbols to ensure consistency in spelling
\newcommand{\topcow}{TopCoW}
\newcommand{\aortaseg}{Aorta24}
\newcommand{\mouseaneurysm}{MouseAneurysm}
\begin{figure}
    \centering
    \includegraphics[width=\textwidth]{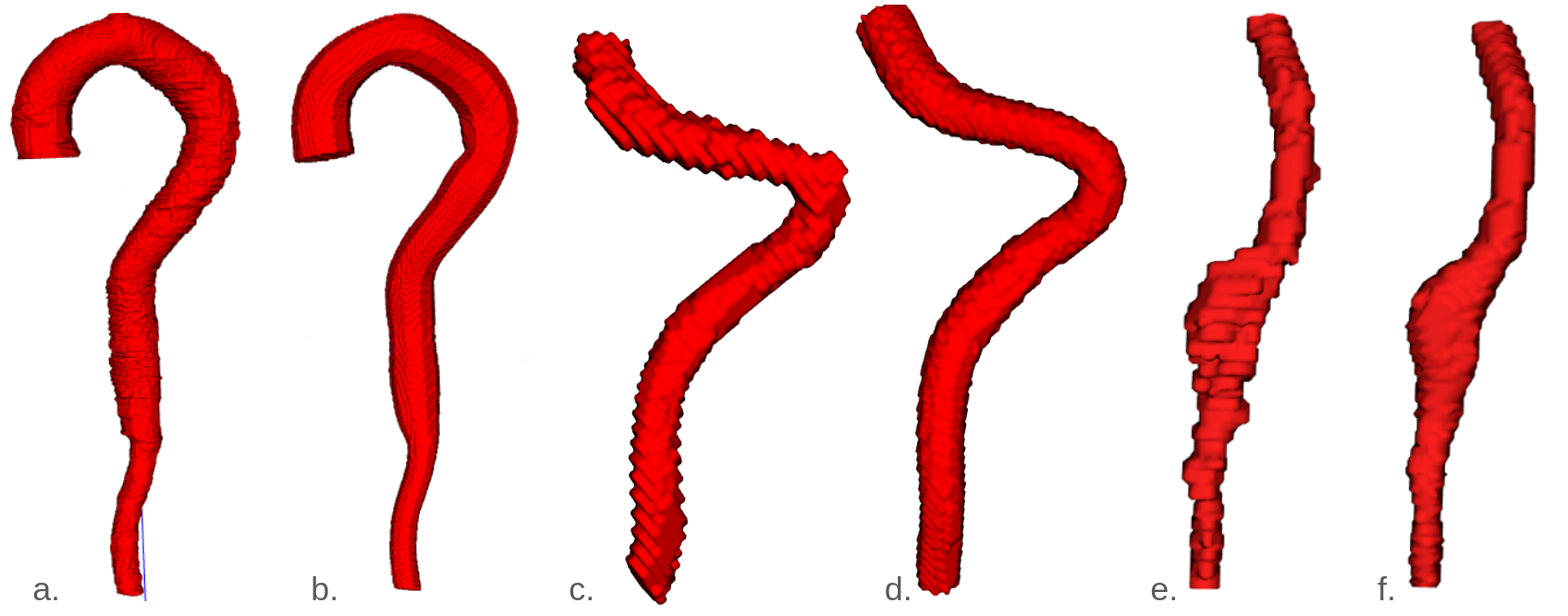}
    \caption{
    Ground truth segmentations (left -- a,c,d) and voxelizations of their corresponding parametric models (right -- b,d,f) are rendered for examples from three datasets: \aortaseg{} (a, b), \topcow{} (c, d), and \mouseaneurysm{} (e, f).
    }\label{fig:qual}%
\end{figure}
\subsubsection{Datasets}
We used three datasets: \aortaseg{}, \topcow{} and \mouseaneurysm{}.
\\
\noindent \emph{\aortaseg{}}~\cite{aortaseg24} consists of 50 3D CT scans of human aortas at an isotropic resolution of 1mm, with manual segmentations.
We merged all the aorta segments under one label and excluded the other vessels.
Centerlines were extracted using Scikit-image skeletonization~\cite{van2014scikit}  and processed using graph-based methods and manual adjustments to extract paths.
The ostia and distal ends were centered using disk fitting in the axial plane, as centerline extraction tools often miss the ends of the vessels.
One sample was excluded due to preprocessing issues.
\\
\noindent \emph{\topcow{}}~\cite{topcowchallenge} is a series of segmentation challenges and datasets of the Circle of Willis.
We obtained access to 50 manual vessel segmentations from the upcoming challenge, which has additional centerlines and radii extracted using  Voreen~\cite{voreen}.
Next, we selected all labels containing only direct paths, resulting in 233 tubular-shaped vessels, which were trained independently.
The Voreen-extracted centerlines also miss the ends of the vessels, which we mitigated by extending the centerline in the direction of the (somewhat noisy) tangents.
For this reason, the endpoints in this dataset are not as accurate as in the other datasets.
We did not have access to the original images for this dataset, so we used the segmentations as input instead.
% We used the segmentations as input instead of images.
The data was upsampled from $0.3\times0.3\times0.6$mm to an isotropic resolution of $0.3$mm.
\\
\noindent \emph{\mouseaneurysm{}} is an in-house dataset of $33$ MRI TOF images of mouse aortas, 25 with aneurysm and 8 control.
The aortas were manually segmented by a non-expert and reviewed by a medical expert.
The centerlines were automatically extracted using the slice-wise center of mass.
The data has a resolution of $0.2\times0.2\times0.5$mm and was upsampled after annotation to $0.2$mm isotropic.
\\
\subsubsection{Implementation details}
We used 3D ResNet18~\cite{feichtenhofer2019slowfast} as the backbone network.
Empirically, we found AdamScheduleFree~\cite{defazio2024road} to converge better than other schedulers.
A tolerance of $0.2$px was used for the centerline loss, which was important for navigating local minima, as the network could focus on optimizing the larger errors.
The number of radial directions was set to $10$ for all datasets.
The number of control points and cross-sections was adjusted to the dataset size; for more details, please refer to the public codebase.
The radial adjustment stage was only used for the \mouseaneurysm{} dataset, as the other datasets have mostly circular cross-sections.
We also experimented with more complex stage schedules, but found no further improvements.
Generally, we found the fitting process relatively robust; for each dataset, it was sufficient to fit the hyperparameters to one sample, and the network would generalize well to the rest of the samples.
We use the Dice score to evaluate the volumetric fit, Hausdorff (tolerance $95\%$), and Chamfer distances to evaluate the centerline fit to the reference centerlines.
The mean and standard deviation of the metrics are reported by averaging the fit of all the individual samples over the entire dataset.
% For computational efficiency, we cropped the data around the vessel of interest. % Mentioned in the methods section
% The voxelization is very slow in its current implementation due to the memory footprint of the gradients, which hinders parallelization. Also mentioned in the methods section.
The implementation codebase and hyperparameters for each dataset are available at:
\href{https://github.com/alinafdima/paravess}{https://github.com/alinafdima/paravess}.
%
% \begin{table}[!htb]
\begin{table}[!tb]
    \centering
    \caption{Overview of the performance of the shape fitting on all the datasets in terms of volumetric fit and proximity to the reference centerline. The average radius is computed as the mean of the radii of the cross-sections and is a characteristic feature of each dataset.}%
    \begin{tabular}{ccccc}
        \toprule
        Dataset & Dice & Average Radius (px) & Cl HD95 & Cl Chamfer\\
        \midrule
        % \hline
        \aortaseg{}      & $94.66 \pm 0.77$ & $16.64 \pm 2.10$ & $4.91 \pm 1.11$ & $2.03 \pm 0.21$ \\
        \topcow{}        & $82.22 \pm 8.77$ & $4.73 \pm 0.98$ & $1.55 \pm 0.69$ & $0.77 \pm 0.22$ \\
        \mouseaneurysm{} & $86.43 \pm 2.71$ & $2.83 \pm 0.27$ & $1.79 \pm 0.25$ & $0.98 \pm 0.08$ \\
        \bottomrule
    \end{tabular}
    \label{tab:datasets}
    \centering
    \caption{Baseline comparison on the \aortaseg{} dataset.
    Our method has desirable parametric properties such as sparsity \emph{without} sacrificing Dice performance.
    % The INR* metrics exclude four subjects with abnormal metrics.
    }%
    \setlength{\tabcolsep}{6pt}
    \begin{tabular}{cccccc}
        \toprule
        Model & Dice $\uparrow$ & Min Angle $\uparrow$ & Aspect ratio $\downarrow$ & Points & Faces\\
        \midrule
        nnUNet & $94.13 \pm 1.79$ & $\mathbf{39.95 \pm 0.10}$ & $\mathbf{1.48} \pm 0.00$ & $76803$ & $153600$ \\
        nnUNet* & $92.43 \pm 1.85$ & $35.51 \pm 0.69$ & $\mathbf{1.73} \pm 0.06$ & $677$ & $1350$ \\
        INR    & $\mathbf{97.50 \pm 0.49}$ & $33.53 \pm 0.38$ & $5.49 \pm 1.85$ & $75814$ & $151625$ \\
        INR*   & $92.58 \pm 1.08$ & $33.55 \pm 0.71$ & $1e+26 $ & $676$ & $1350$ \\
        Ours   & $94.66 \pm 0.77$ & $34.35 \pm 1.89$ & $1.56 \pm 0.09$ & $\mathbf{673}$ & $\mathbf{1326}$ \\
        \bottomrule
    \end{tabular}
    \label{tab:baselines}
\end{table}
\\
\subsubsection{Faithfulness to data}
We first show how well our parametric shape models can fit a variety of vessel shapes.
Qualitative and quantitative results from all datasets are presented in Figure~\ref{fig:qual} and Table~\ref{tab:datasets}.
Due to the near impossibility of obtaining ground truth shape model parameters such as centerlines and radii, we evaluate the faithfulness of the representation using proxy metrics.
Dice score measures the volumetric faithfulness of our fit to the original segmentation, while centerline-based metrics such as Hausdorff and Chamfer distances measure the faithfulness of the centerline.
For a good quality fit, we want our model to deviate from the reference centerline in locations where the reference centerline is incorrect.
Overall, we observe a good fit on all the datasets.
The larger vessels in \aortaseg{} have comparatively larger centerline errors.
The aortas in \emph{\aortaseg{}} have a characteristic curved shape and a significant radius variance due to the many vessels the aorta supplies.
While some aortic dissections and aneurysms may be present, they do not lead to a lot of variance in shape. On this dataset, we can fit well to the data ($94.66\%$ Dice) assuming radial cross-sections, without individual radial adjustments.
On the \emph{\mouseaneurysm{}} dataset, the aneurysms are more challenging to fit due to their non-circular cross-sectional shape and the presence of large aneurysms ($86.63\%$ Dice).
Additional challenges are posed by the time-of-flight modality, where signal loss makes manual segmentation difficult and introduces noise. Interestingly, we found that adjusting the centerline using the segmentation (stage 3) does not improve the quality of the fit for this dataset, possibly due to the irregularities of the shape.
On \emph{\topcow{}}, the vessels are more heterogeneous in their orientation.
The fits are less accurate ($82.22\%$ Dice), owing to a difficulty in procuring high-quality vessel endpoints.
A dedicated branching point detector could greatly help with this.
\\
% \textbf{Quantitative results}.
%
\begin{figure}
    \centering
    \includegraphics[width=\textwidth]{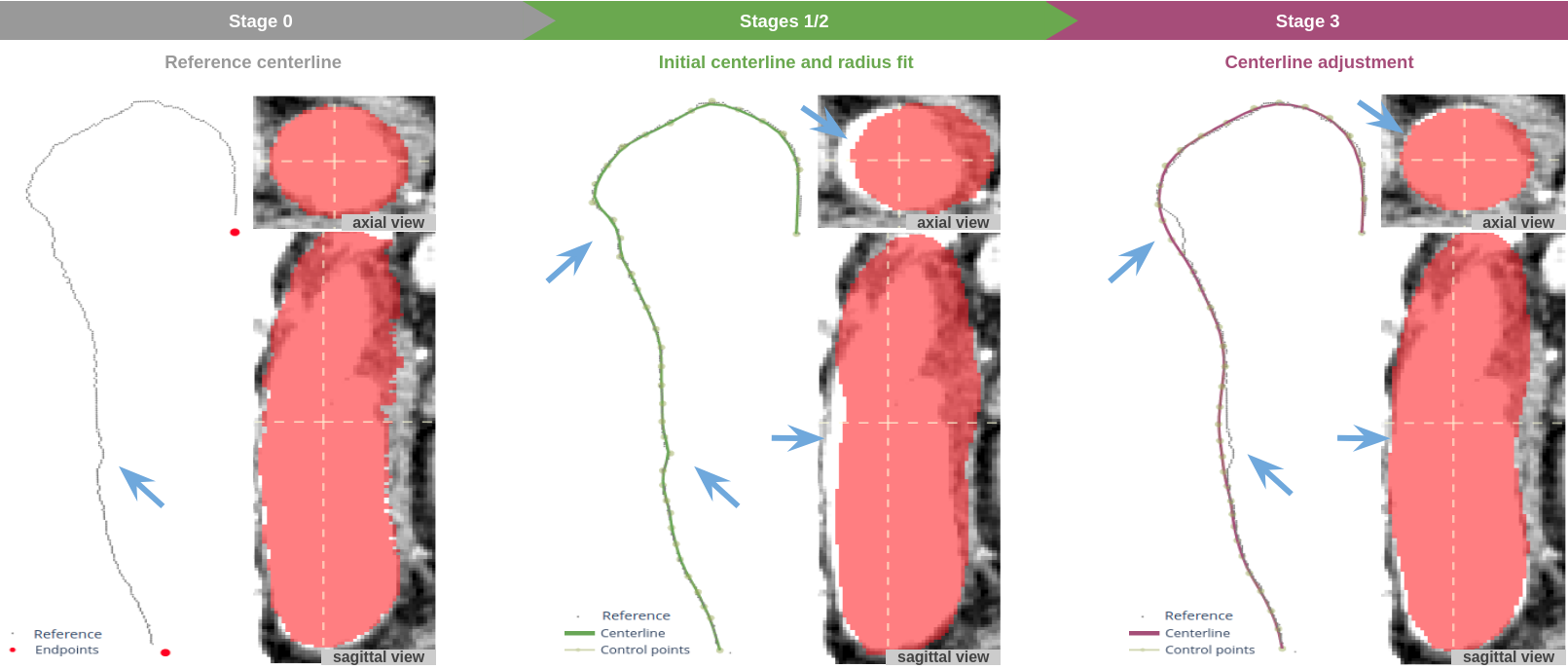}
    \caption{
    A quantitative example of our model's ability to correct the centerline based on the shape.
    In the early stages (green), the centerline fits the reference centerline (gray), which is often noisy.
    The centerline is corrected based on the shape (pink) using the voxelization-based loss.
    This is also reflected in the segmentation mask (red).
    }%
        \label{fig:centerline_correction}
\end{figure}
\\
\subsubsection{Centerline correction based on shape}
In the first stages of training, the parametric model fits the reference centerline due to the centerline distance loss.
This is needed to initialize the radius fit, but the centerline is not yet adjusted to the shape.
During the radius fit, the centerline is fixed, and the radius is adjusted solely based on the segmentation.
Even though the reference centerline is imperfect, the model can estimate the radius correctly due to the regularization effect of the B-spline curves.
During the centerline adjustment stage, the centerline will be further adjusted to the shape without explicit centerline supervision.
This finding highlights the power of the parametric shape fitting of learning explicit parameters from pixels.
Figure~\ref{fig:centerline_correction} shows a qualitative example of this.
\begin{figure}[!htb]%
    \centering
    \includegraphics[width=\textwidth]{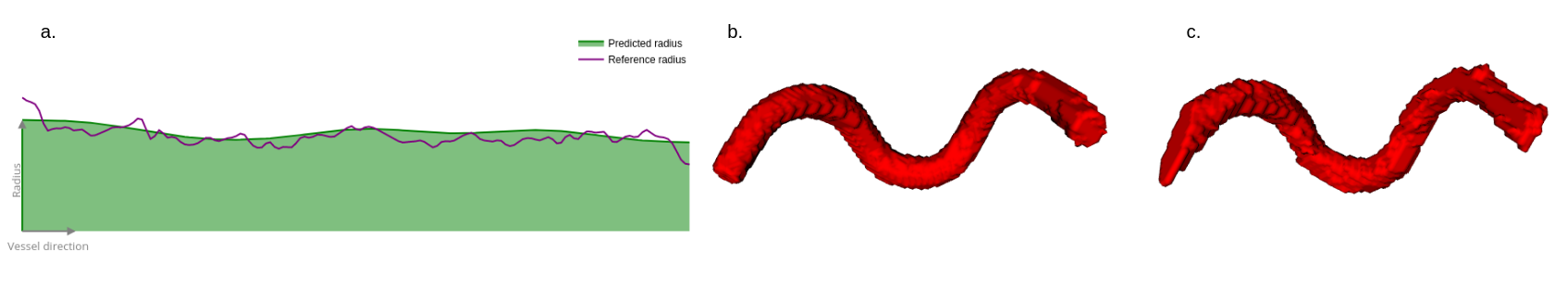}
    \caption{Qualitative results on \topcow, showing the regularization effect of the B-spline curves as radius smoothness.
    Comparison between the reference radius and our radius (a),
    our segmentation (b), reference segmentation (c).}%
    \label{fig:radius}
\end{figure}
\\
\subsubsection{Mesh properties}
Parametric models have an inherent \emph{smoothing capability}, owing to fewer degrees of freedom.
We can see in Figure~\ref{fig:radius} on a typical example on \topcow{} that the predicted radii are much smoother than the reference radii.
% They also offer \emph{full flexibility over the generated meshes}, allowing us to generate higher quality meshes while ensuring quality criteria are met.
In Table~\ref{tab:baselines}, we compare our parametric models to two popular mesh extraction methods: nnUnet~\cite{isensee2024nnu} + marching cubes~\cite{lorensen1998marching} and Implicit Neural Representations(INRs)~\cite{sitzmann2020implicit} + marching cubes.
For both baselines, we also re-meshed the marching cubes output to a target number of faces and vertices comparable to ours (1000 vertices and 2000 faces), marked in the table with a star.
nnUnet was trained on the entire dataset with 5-fold cross-validation, while INRs were fitted per sample, just like our method.
In terms of Dice score, we observe high-quality fits on all three methods (over $94\%$), demonstrating that our parametric models \emph{do not sacrifice fit quality}.
Regarding sparsity, our meshes achieve over \emph{100-fold} reduction in the number of mesh elements.
When we remesh the marching cubes output to a target number of faces and vertices as ours, we observe that the volumetric fit drops to $92\%$ dice on both datasets, indicating that our meshes have \emph{better fitting capabilities at that sparsity level}.
Regarding mesh quality metrics, the sparse meshes we use in training are not better than Marching Cubes but provide computational advantages.
% We also compute mesh quality metrics on the sparse meshes used in training, which are not better than Marching cubes at the moment and should be further investigated.
The parametric representation provides an \emph{interpretable representation}, from which a multitude of meshes can be derived.
% While the quality of the metrics can certainly be improved
Another desirable property of our meshes is the \emph{adjustability} of the parameters and resulting meshes.
This is attractive for surgical planning, where blood flow through a vessel segment can be adjusted to reflect surgical needs.
\\
\subsubsection{Robustness to sparse annotations}
Parametric models can fit even to very sparse annotations, owing to the explicit sparse representation.
To showcase this, we reduced the number of annotation slices in the axial plane on \aortaseg{} 20-fold while, at a fixed distance between slices, and keeping the initial and final slices, and centerline.
With only $5\%$ of the segmented slices, our model exhibits a minor decrease in Dice score of only $1\%$, while significantly reducing the annotation effort.
Furthermore, despite the heavily reduced amount of segmented slices, our model still benefits from the centerline correction stage, observing a further benefit of $0.5\%$ Dice score.
This could help alleviate the cost of vessel annotations, as centerlines are easier to annotate than segmentations.
The results are summarized in Table~\ref{tab:nx}.
\begin{table}[!htb]
    \centering
    \setlength{\tabcolsep}{8pt}
    \caption{Reducing the number of annotations slices in the axial direction 20-fold while keeping the centerline and boundary slices (first and last).}%
    \begin{tabular}{ccccc}
        \toprule
        Slices segmented & Cl correction & Dice & Cl HD95 & Cl Chamfer \\
        \midrule
        $100\%$ & yes & $94.66 \pm 0.77$ & $4.91 \pm 1.11$ & $2.03 \pm 0.21$\\
        $5\%$ & yes & $93.86 \pm 0.93$ & $4.98 \pm 1.07$ & $2.08 \pm 0.22$ \\
        $5\%$ & no & $93.75 \pm 0.94$ & $4.93 \pm 0.92$ & $2.08 \pm 0.22$ \\
        \bottomrule
    \end{tabular}
    \label{tab:nx}
\end{table}

\section{Conclusion}%
\label{sec:conclusion}
In this work, we propose a framework to learn \emph{parametric shape models} for vessels from segmentations via a \emph{differentiable voxelization loss}.
Our approach offers a seamless conversion between parametric, mesh, and voxelized representations for vessels, all within the same differentiable framework.
We show that our approach can extract parametric shape models and generate high-quality meshes, \emph{without sacrificing faithfulness to the data} in terms of volumetric fit.
This paradigm naturally brings many favorable properties and capabilities. 
% We first showed that our model can correct rough centerlines based on the shape and fit highly sparse annotations, while offering flexibility in terms of model parameters and generated mesh.
Its applications range from mesh generation for simulations, sparse annotations, and interpretable vessel analysis.
The vessel shape model can be effectively fitted onto sparse annotations, which can reduce annotation cost. Its parametric nature leads to flexible post-fit manipulation and mesh generation, which is valuable for applications such as surgical planning with intervention simulation and interpretable vessel analysis.

\section{Limitations and Future Work}
\label{sec:limitations}
The flexibility of the parametric model comes with the cost of an increased need for regularization, especially for centerline initialization.
The voxelization loss is slow in its current implementation, but can be improved either through vessel-specific algorithms or more efficient partitioning of the voxel space.
Acknowledging the complexity and difficulty of the task on single vessels, we opted to leave branch handling for multi-vessel modelling in our future work. 
Furthermore, we recognize that our formulation is compatible with the optimization of vessel parameters from images without any reference annotations at inference, which we are currently investigating.
Addressing these limitations opens up the possibility of incorporating state-of-the-art methods from multiple representation spaces within a unified framework, which is our ultimate aim.

\begin{credits}
\subsubsection{\ackname} This study was supported by the following grants:
DFG SFB 1340 "Matrix in Vision" (FKZ: 372486779), "RACOON-PDAC" (FKZ: 01KX2121, funded by NUM) and "PRECISE-MD" (FKZ: 01KD2418A, funded by NUM).
\subsubsection{\discintname}{The authors have no competing interests to declare that are relevant to the content of this article.}
\end{credits}
%
%
%
% ---- Bibliography ----
\bibliographystyle{splncs04}
\bibliography{bibliography}

\begin{thebibliography}{10}
\providecommand{\url}[1]{\texttt{#1}}
\providecommand{\urlprefix}{URL }
\providecommand{\doi}[1]{https://doi.org/#1}

\bibitem{alblas2023implicit}
Alblas, D., Hofman, M., Brune, C., Yeung, K.K., Wolterink, J.M.: Implicit neural representations for modeling of abdominal aortic aneurysm progression. In: International Conference on Functional Imaging and Modeling of the Heart (2023)

\bibitem{bransby2024polycore}
Bransby, K.M., Bajaj, R., Ramasamy, A., {\c{C}}ap, M., Yap, N., Slabaugh, G., Bourantas, C., Zhang, Q.: Polycore: Polygon-based contour refinement for improved intravascular ultrasound segmentation. Computers in Biology and Medicine  \textbf{182},  109162 (2024)

\bibitem{de2022quantification}
de~Carvalho~Macruz, F.B., Lu, C., Strout, J., Takigami, A., Brooks, R., Doyle, S., Yun, M., Buch, V., Hedgire, S., Ghoshhajra, B.: Quantification of the thoracic aorta and detection of aneurysm at {CT}: development and validation of a fully automatic methodology. Radiology: Artificial Intelligence  (2022)

\bibitem{defazio2024road}
Defazio, A., Yang, X., Mehta, H., Mishchenko, K., Khaled, A., Cutkosky, A.: The road less scheduled. Advances in Neural Information Processing Systems  (2024)

\bibitem{voreen}
Drees, D., Scherzinger, A., H{\"a}gerling, R., Kiefer, F., Jiang, X.: Scalable robust graph and feature extraction for arbitrary vessel networks in large volumetric datasets. BMC bioinformatics  (2021)

\bibitem{feichtenhofer2019slowfast}
Feichtenhofer, C., Fan, H., Malik, J., He, K.: Slowfast networks for video recognition. In: Proceedings of the IEEE/CVF international conference on computer vision. pp. 6202--6211 (2019)

\bibitem{franklinpnpoly}
Franklin, W.R.: Pnpoly-point inclusion in polygon test. Web site: http://www. ecse. rpi. edu/Homepages/wrf/Research/Short\_Notes/pnpoly. html  (2006)

\bibitem{aortaseg24}
Imran, M., Krebs, J.R., Sivaraman, V.B., Zhang, T., Kumar, A., Ueland, W.R., Fassler, M.J., Huang, J., Sun, X., Wang, L., et~al.: Multi-class segmentation of aortic branches and zones in computed tomography angiography: The aortaseg24 challenge. arXiv preprint arXiv:2502.05330  (2025)

\bibitem{isensee2024nnu}
Isensee, F., Wald, T., Ulrich, C., Baumgartner, M., Roy, S., Maier-Hein, K., Jaeger, P.F.: nnu-net revisited: A call for rigorous validation in 3d medical image segmentation. In: International Conference on Medical Image Computing and Computer-Assisted Intervention (2024)

\bibitem{lazarow2022instance}
Lazarow, J., Xu, W., Tu, Z.: Instance segmentation with mask-supervised polygonal boundary transformers. In: Proceedings of the IEEE/CVF Conference on Computer Vision and Pattern Recognition (2022)

\bibitem{lorensen1998marching}
Lorensen, W.E., Cline, H.E.: Marching cubes: {A} high resolution 3d surface construction algorithm. In: Proceedings of the 14th Annual Conference on Computer Graphics and Interactive Techniques, {SIGGRAPH} (1987)

\bibitem{meng2023deepmesh}
Meng, Q., Bai, W., O’Regan, D.P., Rueckert, D.: Deepmesh: mesh-based cardiac motion tracking using deep learning. IEEE transactions on medical imaging  (2023)

\bibitem{menten2023skeletonization}
Menten, M.J., Paetzold, J.C., Zimmer, V.A., Shit, S., Ezhov, I., Holland, R., Probst, M., Schnabel, J.A., Rueckert, D.: A skeletonization algorithm for gradient-based optimization. In: Proceedings of the IEEE/CVF International Conference on Computer Vision (2023)

\bibitem{numata2016blood}
Numata, S., Itatani, K., Kanda, K., Doi, K., Yamazaki, S., Morimoto, K., Manabe, K., Ikemoto, K., Yaku, H.: Blood flow analysis of the aortic arch using computational fluid dynamics. European Journal of Cardio-Thoracic Surgery  (2016)

\bibitem{rist2023flexible}
Rist, L., Taubmann, O., Ditt, H., S{\"u}hling, M., Maier, A.: Flexible unfolding of circular structures for rendering textbook-style cerebrovascular maps. In: International Conference on Medical Image Computing and Computer-Assisted Intervention (2023)

\bibitem{rogers2000introduction}
Rogers, D.F.: An introduction to NURBS: with historical perspective. Elsevier (2000)

\bibitem{schwarz2010fast}
Schwarz, M., Seidel, H.P.: Fast parallel surface and solid voxelization on gpus. ACM transactions on graphics (TOG)  (2010)

\bibitem{sitzmann2020implicit}
Sitzmann, V., Martel, J., Bergman, A., Lindell, D., Wetzstein, G.: Implicit neural representations with periodic activation functions. Advances in neural information processing systems  \textbf{33},  7462--7473 (2020)

\bibitem{timmins2022deep}
Timmins, K.M., van~der Schaaf, I.C., Vos, I., Ruigrok, Y.M., Velthuis, B.K., Kuijf, H.J.: Deep learning with vessel surface meshes for intracranial aneurysm detection. In: Medical Imaging 2022: Computer-Aided Diagnosis. SPIE (2022)

\bibitem{van2014scikit}
Van~der Walt, S., Sch{\"o}nberger, J.L., Nunez-Iglesias, J., Boulogne, F., Warner, J.D., Yager, N., Gouillart, E., Yu, T.: Scikit-image: Image processing in python. PeerJ  \textbf{2}, ~e453 (2014)

\bibitem{topcowchallenge}
Yang, K., Musio, F., Ma, Y., Juchler, N., Paetzold, J.C., Al-Maskari, R., Höher, L., Li, H.B., Hamamci, I.E., Sekuboyina, A., et~al.: Benchmarking the {CoW} with the {TopCoW} challenge: Topology-aware anatomical segmentation of the circle of willis for {CTA} and {MRA} (2024), \url{https://arxiv.org/abs/2312.17670}

\end{thebibliography}
\end{document}